\pdfoutput=1

\documentclass[11pt]{article}

\usepackage{acl}

\usepackage{times}
\usepackage{latexsym}
\usepackage{xcolor}

\usepackage[T1]{fontenc}

\usepackage[utf8]{inputenc}

\usepackage{microtype}

\usepackage{inconsolata}

\usepackage{tabularray}
\usepackage{enumitem}
\usepackage{graphicx}
\usepackage{amsmath}
\usepackage{amssymb}
\usepackage{booktabs}
\usepackage{tabularx}
\usepackage{authblk} 
\usepackage{color, booktabs,subcaption,amsfonts,dcolumn}
\usepackage{amsfonts}
\usepackage{graphicx}

\usepackage{colortbl}
\usepackage{multirow,diagbox,makecell}
\usepackage{tikz}

\usepackage{pifont}
\usepackage{booktabs}

\usepackage{capt-of}
\title{ViCor: Bridging Visual Understanding and Commonsense Reasoning with Large Language Models}

\author[1]{Kaiwen Zhou}
\author[2]{Kwonjoon Lee}
\author[2]{Teruhisa Misu}
\author[1]{Xin Eric Wang}
\affil[1]{University of California, Santa Cruz}
\affil[2]{Honda Research Institute}

\begin{document}

\maketitle
\begin{abstract}

In our work, we explore the synergistic capabilities of pre-trained vision-and-language models (VLMs) and large language models (LLMs) on visual commonsense reasoning (VCR) problems. 
We find that VLMs and LLMs-based decision pipelines are good at different kinds of VCR problems. Pre-trained VLMs exhibit strong performance for problems involving understanding the literal visual content, which we noted as visual commonsense understanding (VCU). For problems where the goal is to infer conclusions beyond image content, which we noted as visual commonsense inference (VCI), VLMs face difficulties, while LLMs, given sufficient visual evidence, can use commonsense to infer the answer well. 
We empirically validate this by letting LLMs classify VCR problems into these two categories and show the significant difference between VLM and LLM with image caption decision pipelines on two subproblems. 
Moreover, we identify a challenge with VLMs' \textit{passive} perception, which may miss crucial context information, leading to incorrect reasoning by LLMs. 
Based on these, we suggest a collaborative approach, named \textbf{ViCor}, where pre-trained LLMs serve as problem classifiers to analyze the problem category, then either use VLMs to answer the question directly or \textit{actively} instruct VLMs to concentrate on and gather relevant visual elements to support potential commonsense inferences. 
We evaluate our framework on two VCR benchmark datasets and outperform all other methods that do not require in-domain fine-tuning.
\end{abstract}




\section{Introduction}

The problem of visual commonsense reasoning (VCR)~\citep{zellers2019vcr, hesselhwang2022abduction, schwenk2022okvqa} expands upon the traditional visual question answering~\citep{VQA, balanced_vqa_v2}. VCR requires machines to understand complex visual scenes, extract crucial visual content, and utilize commonsense knowledge for drawing novel conclusions that go beyond the explicit information present in the image. Previous methods have utilized pre-trained large language models and pre-trained or fine-tuned vision-language models~\citep{hu2022promptcap,shao2023prompting,you2023idealgpt} to solve VCR problems in few-shot or fine-tuned setting.

However, some open questions exist on how VLMs and LLMs can efficiently and effectively collaborate to solve these VCR problems. Firstly, what roles do LLMs and VLMs play in solving VCR problems with their different capabilities? Secondly, how do we maximize their capabilities to solve the VCR problems without in-domain fine-tuning? 

\begin{figure*}[t]
\vspace{-7mm}
  \centering
  \includegraphics[width=1\textwidth]{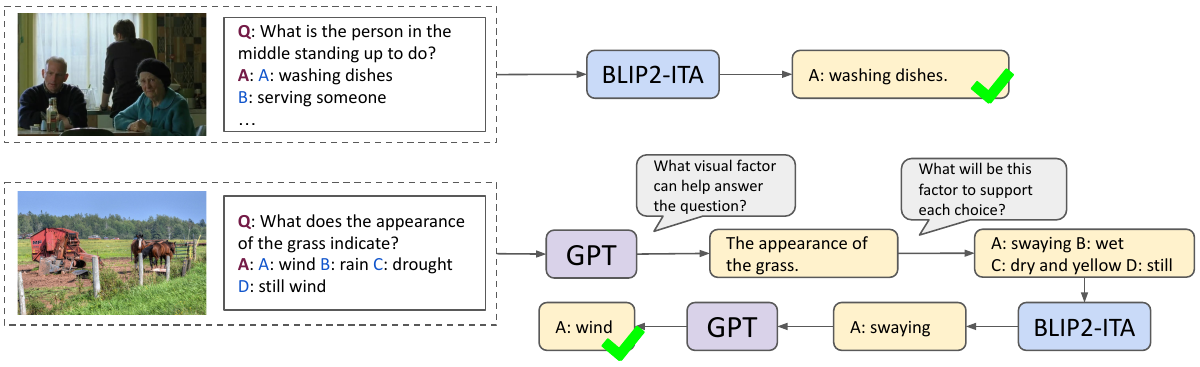}
  \caption{Two examples demonstrating different kinds of visual commonsense reasonings require different model capabilities. \textbf{Upper: Visual commonsense understanding (VCU)} requires the model to understand high-level concepts and attributes such as actions, events, relations, \textit{etc}, which pre-trained VLMs can achieve via image-text alignment (ITA). \textbf{Lower: Visual commonsense inference (VCI)} requires the model to generate conclusions or explanations based on input image. Overlooking visual clues can result in erroneous conclusions. LLMs steer VLMs in discovering vital visual cues for answer support. The LLM employs the top ITA-scored visual clue (e.g.,``It is cloudy.") to perform commonsense inference.}
  \label{fig:1}
  \vspace{-3mm}
\end{figure*}

To answer these two questions, as shown in Figure~\ref{fig:1}, we first find that VLMs themselves can solve the problems requiring the model to recognize various low-level visual patterns and understand high-level concepts like actions, events, and relations indicated by those visual patterns. In the meanwhile, solving problems that require the model to deduce conclusions or form explanations based on visual observation relies more on the commonsense reasoning capabilities of LLMs. This kind of problem requires a broad array of commonsense knowledge about the world, including cause-and-effect relationships, intentions, and mental states~\citep{sap-etal-2020-commonsense}. 
To validate this finding, we first note these two kinds of VCR problems as \textit{visual commonsense understanding (VCU)} and \textit{visual commonsense inference (VCI)}. Then, we instruct LLMs to classify VCR question into these two categories. We empirically find that, for VCU problems, VLMs like BLIP2 can achieve better results than LLM+caption pipeline~\cite{yang2021empirical} with their visual understanding capabilities, and for VCI problems, the LLM+caption pipeline is better. 

In the meanwhile, we observe that image captions provided by VLMs often lack crucial contextual information necessary for answering questions. This poses a particular challenge for commonsense inference problems, as inferences are often defeasible given additional context~\citep{10.1162/daed_a_01906}. To illustrate this issue, consider the example depicted in Figure~\ref{fig:1} (bottom). 
At first glance, it may appear that there's nothing noteworthy beyond horses on a grassy farm, leading one to select ``D: still wind" as an answer. However, upon closer examination of the swaying grass, we must revise our conclusion to ``A: wind."  Existing perception modules, including VLMs, operate in a feed-forward manner and cannot adjust their perception based on a high-level understanding or inference. 
To address this, we propose instructing LLMs to intervene with VLMs in cases where they are uncertain about inference, indicating a lack of sufficient visual evidence. This intervention would guide VLMs to focus on specific \textit{visual factors}, such as weather or emotions, to support commonsense inferences. 


Based on these findings, we propose the \textbf{ViCor} framework, which employs the following components: (1) LLMs functioning as problem type classifiers (VCU and VCI), VLM commanders for directing VLMs based on problem classification, and visual commonsense reasoners to harness their extensive world knowledge and reasoning capabilities. (2) Pre-trained VLMs are responsible for visual recognition and understanding. Communication between LLMs and VLMs occurs through text, such as image captions, as they are universal medium for all existing models. On VCR~\citep{zellers2019vcr} and A-OKVQA~\citep{schwenk2022okvqa}, our method achieves state-of-the-art results among methods \textit{without} supervised in-domain fine-tuning. 

\section{Related Work}

\paragraph{Visual Commonsense Reasoning~} 
Visual Commonsense Reasoning (VCR)~\cite{zellers2019vcr,hesselhwang2022abduction,schwenk2022okvqa} is an emerging research area that aims to endow AI models with a human-like understanding and reasoning of visual scenes.
The goal is to understand high-level concepts such as events, relations, and actions and infer unobservable aspects such as intents, causal relationships, and future actions, requiring the integration of visual understanding, and commonsense knowledge. The VCR task was introduced by ~\citet{zellers2019vcr}, where models must answer a question about an image, given a set of four possible answers. Further, more datasets focus on more types of reasoning were proposed~\cite{park2020visualcomet,hesselhwang2022abduction,schwenk2022okvqa}. 
Most methods treat VCR as an image-text alignment problem, where they encode the commonsesense inference and the visual input, then predict the alignment score of the image-text pair via a classification head or image-text similarity~\cite{zellers2019vcr,chen2020uniter,zellers2022merlotreserve,hesselhwang2022abduction}. Although achieving impressive performance, the generalizability of these methods is limited by supervised training. Recently, several works have leveraged large language models for visual commonsense reasoning~\cite{hu2022promptcap,shao2023prompting,you2023idealgpt}. However, \cite{hu2022promptcap,shao2023prompting} require some VLMs trained on the datasets to provide visual information. \cite{you2023idealgpt} use LLMs to decompose the main problem and use VQA models to acquire visual information.
Our work systematically studies the strength of pre-trained VLMs and the reasoning abilities of LLMs on visual commonsense reasoning problems and proposes a framework to efficiently and effectively leverage the advantage of both models. 

\vspace{0.2ex}
\paragraph{Large Language Models for Vision-and-Language Tasks~} Benefiting from the rich knowledge in LLMs, they have been used for various vision-and-language tasks in a zero-shot or few-shot manner. ~\citet{yang2021empirical,hu2022promptcap,shao2023prompting} leverage LLMs for OK-VQA task~\cite{okvqa} by feeding the caption, question, candidate answers by VQA models, etc. to GPT3 models, and prompt the GPT model to answer the question with its pre-trained knowledge. ~\citet{wang2022language} propose to use LLMs with image descriptors for video-language tasks. More recently, with the discovery of LLMs' tool using ability~\cite{yao2023react,Schick2023ToolformerLM}, LLMs were equipped with various visual tools~\cite{Gupta_2023_CVPR,surismenon2023vipergpt,shen2023hugginggpt, lu2023chameleon,wu2023visual} and achieved significant performance in Compositional Visual Question Answering, Science Question Answering tasks~\cite{suhr2018corpus,hudson2019gqa,lu2022learn}. 
Different from these works, we study a more complex and challenging task with different levels of reasoning, including requiring reasoning beyond direct image observation. In our method, we maximize the capabilities of LLMs to perform reasoning for problem classification, visual information query, and commonsense reasoning.

\section{Visual Commonsense Reasoning}

\subsection{Problem Categorization} 
We first illustrate our categorization of VCR problems to distinguish the capabilities of VLMs and LLMs. 
\paragraph{Visual Commonsense Understanding} The visual commonsense understanding (VCU) problem requires the model to judge if a text $T$ describing a concept or an attribute aligns with the image $I$:
\begin{equation}
    e = F(I, T)
\end{equation}
where $e$ stands for evaluation of $T$ by model $F$. To answer these questions, the model needs to be able to map the low-level visual observations, such as objects and spatial relations to various high-level visual concepts and attributes, such as landmarks, actions, events, and relations.

\paragraph{Visual Commonsense Inference} The visual commonsense inference (VCI) problem usually requires the model to evaluate the plausibility of an inference about the image. 
Besides understanding the literal content in the image as in VCU, evaluating the inferences $T$ in VCI problems needs involves drawing novel conclusions or explanations from these visuals, often using (non-visual) commonsense knowledge, based on some visual observations $\{o_i\}$ derived from the image:
\begin{equation} \label{eq:2}
    e = F(\{o_{i}\}, T)
\end{equation}
Here, $o_{i}$ could be some visual observations or high-level visual commonsense understanding. 
Examples of non-visual commonsense knowledge could be the purpose of an object, people's opinions about an object, potential future events, etc. 
We show the performance difference between VLMs and LLMs-based decision models on two sub-problems in Table~\ref{tab:ablations}, which will be illustrated in Sec.~\ref{sec:6.1}.

\subsection{Problem Formulation}

 Both categories of visual commonsense reasoning tasks share a common formulation. In visual commonsense reasoning, the input consists of two parts: an image denoted as $I$ and a multiple-choice question input represented as ${q, {c_i}}$, where $q$ corresponds to the question, and $c_i$ stands for the $i$-th answer choice. The model needs to choose the choice $c_i$ that is most likely to be true based on the image $I$.

\section{The ViCor Framework} 

\begin{figure*}[t]
    
  \centering
  \includegraphics[width=1\textwidth]{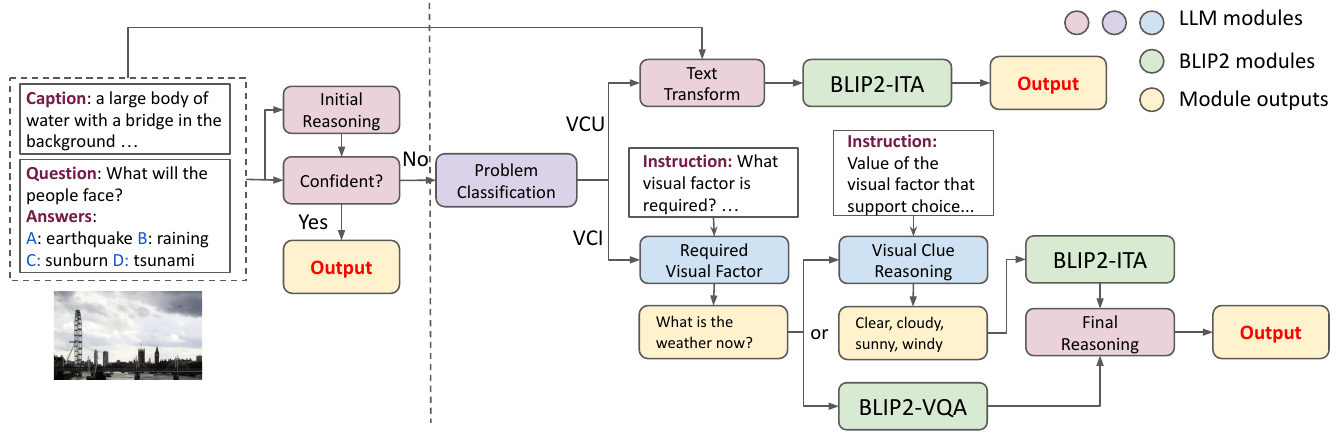}
  \caption{\textbf{Our ViCor framework.} Given a visual commonsense reasoning problem and a caption, our framework will leverage LLM to perform initial reasoning and confidence check. If the reasoning is not confident, the LLM will perform problem classification and acquire visual information according to the problem type. $*$Note that the final reasoning takes the question and the caption as input as well.} 
  
  \label{fig:2}
  \vspace{-4mm}
\end{figure*}

To enable more effective collaboration between LLMs and VLMs, as shown in Figure~\ref{fig:2}, we design an approach that involves a multi-step process. First, a pre-trained LLM takes the initial perception result (\textit{i.e.}, image caption), a question-answer pair, and instructions as input to evaluate potential answer candidates. 
Then, if the LLM is not confident about its reasoning, 
it will select to use VLMs to directly answer the question or to guide the VLMs to collect target visual information for re-evaluation.


\subsection{Large Language Models as VCR Reasoner}
Evaluating answer choices in VCR requires drawing new conclusions based on commonsense knowledge, which LLMs excels at~\citep{anil2023palm}. On the other hand, pre-trained vision-and-language models have exhibited a capability for visual understanding, such as image captioning and image-text alignment, with a demonstrated ability to generalize across various datasets~\citep{li2023blip2}. Therefore, we decided to harness the strengths of vision-and-language models for visual understanding and the capabilities of large language models for evaluating answer candidates in the context of visual commonsense reasoning. 

Captioning serves as a fundamental unsupervised pre-training task and the most generalized capabilities of pre-trained VLMs, which capture the most salient information from an image. 
Therefore, we first prompt the LLMs to take the caption of the image $C_I$ as the initial information and perform chain-of-thought reasoning on the question:
\begin{equation}
    r_1 = LLM(\{c_i\}, q, C_I).
\end{equation}
The reasoning result $r_1$ includes both intermediate reasoning steps and the final answer. However, it's important to note that the image caption may not encompass all the relevant information within the image, potentially omitting critical contextual details essential for answering the question. In such cases, it becomes necessary to gather additional relevant visual observations from the image. 
Before this, we must first judge whether there is a lack of supportive visual evidence that would allow us to make a confident decision. As in Figure~\ref{fig:2}, we let the LLM take the initial reasoning $r_1$ and the history prompt as input to judge if current visual information adequately supports the decision. If it does, the model will directly output the result. Conversely, if there is a lack of sufficient evidence, the model will progress to the second stage, where it will actively seek additional visual evidence.

\subsection{Large Language Models as VCR Problem Classifier}
\begin{figure*}[t]

  \centering
  \includegraphics[width=1\textwidth]{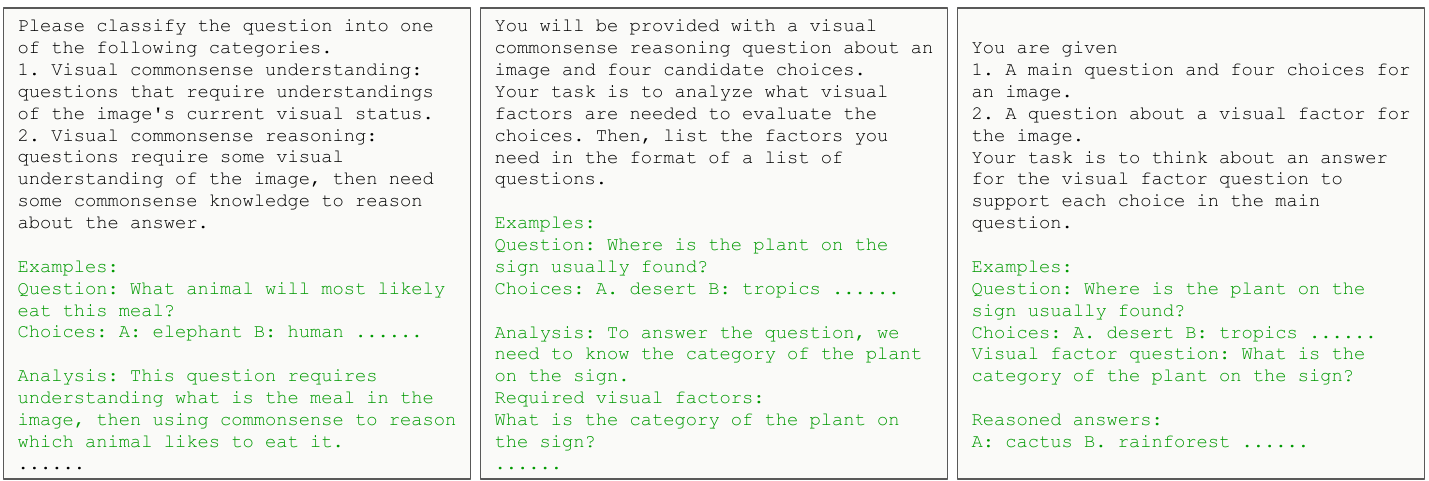}
  \vspace{-5mm}
  \caption{Three simplified prompt examples demonstrating how we define prompts to classify the problem (\textbf{left}), reason visual factors (\textbf{middle}), and think about visual observations regarding visual factors (\textbf{right}).}
  \label{fig:3}
  \vspace{-5mm}
\end{figure*}

As mentioned before, we found the capabilities of VLMs are suitable for solving VCU problems, and LLMs are more capable of solving VCI problems. Therefore, we propose to leverage VLMs in distinct manners when facing different problem types. 
To this end, we first prompt the LLM to classify the problem into two categories. To achieve this, we provide the definitions of these two categories in the prompt. Additionally, we include a set of manually annotated in-context examples to aid in problem classification, where the questions of in-context examples are selected from the training set. Figure~\ref{fig:3} illustrates the prompt.

\subsection{Large Language Models as VLM Commander} \label{sec:4.3}
The pre-training dataset of vision-and-language models contains millions to billions of image-text pairs. Therefore, VLMs have learned the mapping between visual features and the high-level commonsense concept well. 
In light of this, for \textit{visual commonsense understanding (VCU)} problems, we leverage pre-trained VLM in a zero-shot manner. Specifically, for each choice $c_i$, we first instruct the LLM to transfer it and the question to a declarative sentence with instruction and in-context examples:
\begin{equation}
    s_i = LLM(q, c_i)
\end{equation}
For instance, for the question \texttt{What will the people face?} and the choice \texttt{earthquake}, we will transform them to \texttt{The people will face earthquake}.  Then, we feed $s_i$ and the image $I$ to the pre-trained VLM to calculate the image-text alignment score. Following ~\citep{li2023blip2}, we use the sum of ITM and ITC scores to compare choices:
\begin{equation}
    S_i = ITM(I, s_i) + ITC(I, s_i)
    \label{eq:5}
\end{equation}
We will directly take the choice with the highest score as the final output.

For the \textit{visual commonsense inference (VCI)} problems, the model needs to acquire related visual observations and use relevant commonsense knowledge to reason about the answer. Some crucial visual observations are often neglected in the descriptions of the image. 
Therefore, as in Figures~\ref{fig:2} and~\ref{fig:3}, we first prompt the LLMs to think about some \textit{visual factors} $f_j$ that influence the answer to the question, like `the action of the person', `the interaction between people', etc. Then, we could acquire the visual observation of the visual factor in the image with a visual question-answering model by asking a question about the visual factor:
\begin{equation}
    o_j = VQA(I, f_j)
\end{equation}
where $o_j$ is the answer to the question which we call \textit{visual clue}.
However, the answer of VQA does not consider the context of the main question and therefore may lack the most related information. 
To better leverage the contextualized reasoning capabilities of LLMs, we further propose to prompt the LLM to reason the potential instantiations of the visual factors that can support the choices as in Figure~\ref{fig:3}:
\begin{equation}
    o_{ij} = LLM(f_j, c_i, q)
    \label{eq:7}
\end{equation}
For instance, when $f_j$ is ``category of the plant," the potential values for $o_{ij}$ may include specific plant names like ``cactus." Then, we could leverage the image-text matching (ITM) and image-text contrastive (ITC) functions of pre-trained VLMs to select the observation that most align with the image among the observations for each choice $i$:
\begin{align}
    o_{j} &= o_{kj} \\ 
    \text{ where } k&=argmax_{i}\{ITM(o_{ij}, I) + ITC(o_{ij}, I)\}
\end{align}
Finally, we append the \textit{visual clues}  $\{o_{j}\}$ after the caption as extra information for LLM to perform final reasoning:
\begin{equation}
    r_2 = LLM(\{c_i\}, q, C_I, \{o_{j}\})
\end{equation}

\section{Experiments}

\subsection{Datasets} \label{sec:5.1}
We mainly evaluate our approach on two datasets focused on visual commonsense reasoning: VCR~\citep{zellers2019vcr} and AOKVQA~\citep{schwenk2022okvqa}.\footnote{We provide the result on OKVQA dataset in Appendix.~\ref{appendix:a}.} Both datasets formulate visual commonsense reasoning as 4-choice QA problems about an image, containing various visual commonsense understanding and inference problems. VCR dataset focuses on human-centric visual commonsense reasoning problems. In contrast, A-OKVQA dataset requires various commonsense knowledge about common objects and events in daily life. For A-OKVQA, we use the validation set with 1145 examples. For VCR dataset, we randomly sample 3000 / 26534 examples from the validation set for the ablation study, and sample 500 examples to compare with other methods due to the GPT4 API cost. We divide the image from left to right into three bins and name the person depending on which bin they are located in when feeding text to VLMs and LLMs, similar to~\citep{you2023idealgpt}. The performance of both datasets is evaluated by accuracy.  

\begin{table*}[t]
\vspace{-10mm}
\centering
\setlength{\abovecaptionskip}{8pt}
\setlength{\belowcaptionskip}{8pt}
\caption{Ablations on the effect of LLMs and VLMs on VCR~\cite{zellers2019vcr} and A-OKVQA~\cite{schwenk2022okvqa} datasets. We use GPT-3.5-turbo-0613 for LLM-based methods. *Orig means using the declarative sentences transformed by LLM (Eq.\ref{eq:5}). *Clue means using the clues generated by LLM for image-text alignment (Eq.\ref{eq:10}). All numbers indicate accuracy (\%). ``Conf" indicates the samples where the LLM-Caption baseline shows confidence in its initial reasoning, while ``!Conf" indicates cases where it lacks confidence.}
\resizebox{0.92\textwidth}{!}{
\begin{tabular}{cccccccccc}
\toprule
\multirow{3}{*}{\textbf{Decision Model}}& \multirow{3}{*}{\textbf{Visual Info}}  & \multicolumn{4}{c}{\textbf{AOKVQA}} & \multicolumn{4}{c}{\textbf{VCR}} \\
     &  & \multicolumn{2}{c}{\textbf{VCU}} & \multicolumn{2}{c}{\textbf{VCI}} & \multicolumn{2}{c}{\textbf{VCU}} & \multicolumn{2}{c}{\textbf{VCI}}\\
    &   & Conf & !Conf & Conf & !Conf & Conf & !Conf & Conf & !Conf \\
    \midrule 
 \multirow{2}{*}{BLIP2-Pretrain} &Orig* & 76.5 & 66.3 & 56.5 & 50.9 & 70.0 & 56.3 & 59.2 & 47.4 \\
 & LLM Clue*& 74.4 & 63.0 & 60.2 & 56.1 & 70.6 & 56.7 & 63.3 & 49.2   \\ 
  \midrule
 \multirow{3}{*}{LLM} & Caption &78.9 &55.1 & 85.2 & 50.9 & 75.3 & 46.6 & 65.3 & 41.9  \\
& Caption + VQA Clue &77.5 & 56.2 & 82.4 &  54.9 & 75.9 & 51.9 & 65.3 & 47.3 \\
& Caption + LLM Clue & 79.2 & 65.6 & 81.5 &  64.2 &  72.9 & 58.1  & 57.1 & 52.9 \\
  \midrule
\multicolumn{2}{c}{Num. of Examples} & 289 & 575 & 108 & 173 & 170 & 1779 & 49 & 1002\\
\bottomrule
\end{tabular}
}
\vspace{-3mm}
\label{tab:ablations}
\end{table*}

\subsection{Implementation Details and Baselines}
In our experiments, we use GPT-3.5-turbo-0613 and GPT-4-0613 as the LLMs for reasoning. To ensure reproducibility, we set the temperature of the LLMs to 0. For image captioning, we employ LLAVA-7B-v1.1. Furthermore, we use the pre-trained BLIP2 model for image-text alignment and BLIP2-FlanT5 XL for visual question answering. The number of in-context examples used in the prompts shown in Figure~\ref{fig:3} is 6, 1, and 3, respectively. All the questions in the in-context examples are from the training set.

We implement the following training-free baselines for comparison:
\textbf{ (1) BLIP2-Pretrain}~\citep{li2023blip2}: We use the pre-trained BLIP-2 model directly to perform image-text alignment on both datasets. On both datasets, we utilize GPT-3.5-turbo-0613 to transform the questions and choices into declarative sentences and feed them to the BLIP-2 model to calculate the image-text alignment score. We select the choice with the highest alignment score as the answer.
\textbf{(2) IdealGPT}~\citep{you2023idealgpt}: It prompts LLMs to break down the question and iteratively query a VQA model to answer sub-questions for visual reasoning. In our experiments, we employ the original source code of IdealGPT while utilizing the \textit{same} version of LLM and VLMs for caption, VQA, and reasoning as our method. 

\vspace{-2mm}
\section{Results and Analysis}

\begin{table*}[t]
    \vspace{-10mm}
    \centering
    \setlength{\abovecaptionskip}{8pt}
    
\setlength{\belowcaptionskip}{8pt}
\begin{minipage}{0.45\textwidth} 
\centering
\caption{Comparison between ViCor and other methods on VCR Q$\rightarrow$A task. * Results on full validation set. $\dag$ CoT indicates the same setting as `Caption' baseline in Table.~\ref{tab:ablations}: given caption and perform chain-of-thought reasoning.} 
\resizebox{1\textwidth}{!}{
        \begin{tabular}{clc}
            \toprule
             & Method & Acc.(\%) \\
            \midrule
            \multirow{2}{*}{\rotatebox[origin=c]{90}{Sup.}} & R2C~\cite{zellers2019vcr} & 67.3 \\
            & *MERLOT~\cite{zellers2021merlot} & 79.4 \\
            \midrule 
            \multirow{8}{*}{\rotatebox[origin=c]{90}{ICL}} & BLIP2-Pretrain~\cite{li2023blip2} & 51.2 \\
             & \cellcolor[gray]{0.9} \textit{GPT-3.5} & \cellcolor[gray]{0.9}\\
             & $^{\dag}$CoT  & 43.8 \\
             & IdealGPT~\cite{you2023idealgpt} & 47.9 \\
             & ViCor (ours) & 55.4 \\
             & \cellcolor[gray]{0.9}\textit{ GPT-4} &\cellcolor[gray]{0.9} \\
             & CoT &  57.8 \\
             & ViCor (ours) & 59.8 \\
            \bottomrule
        \end{tabular}
    }
    \label{tab:main vcr}
\end{minipage}
\hfill
\begin{minipage}{0.45\textwidth} 
\centering
\caption{Comparison between ViCor and other methods on A-OKVQA dataset. *Both PromptCap and Prophet trained VLMs on A-OKVQA dataset as part of the module. Sup. indicates supervised methods, and ICL means methods using in-context learning.}
\resizebox{0.95\textwidth}{!}{
        \begin{tabular}{clc}
            \toprule
             & Method & Acc.(\%)\\
            \midrule
            \multirow{4}{*}{\rotatebox[origin=c]{90}{Sup.}} & GPV-2~\cite{kamath2022webly} & 60.3 \\
            & *PromptCap~\cite{hu2022promptcap} & 73.2 \\ 
            & *Prophet~\cite{shao2023prompting} & 76.4 \\
             & InstructBLIP~\cite{instructblip} & 81.0 \\
            \midrule
            \multirow{8}{*}{\rotatebox[origin=c]{90}{ICL}} & BLIP2-Pretrain~\cite{li2023blip2} & 65.6 \\
             & \cellcolor[gray]{0.9} \textit{GPT-3.5 }&\cellcolor[gray]{0.9}\\
             & CoT &  63.3 \\
             & ViCor (ours) &  70.9 \\
             & \cellcolor[gray]{0.9}\textit{ GPT-4} & \cellcolor[gray]{0.9}\\
             & CoT &  70.3 \\
             & AssistGPT~\cite{gao2023assistgpt} &  74.7 \\
             & ViCor (ours) & 75.6 \\
            \bottomrule
        \end{tabular}
        }
    \label{tab: main sherlock}
\end{minipage}
\vspace{-6mm}
\end{table*}

\subsection{Ablation Study} ~\label{sec:6.1}
\vspace{-2mm}

We conduct ablation studies about VLMs and LLMs collaboration on VCR and AOKVQA datasets. Results are shown in Table~\ref{tab:ablations}.

\vspace{1mm}
\noindent \textbf{How do VLM and LLM compare on visual commonsense reasoning?} By comparing the first row and the third row in Table~\ref{tab:ablations}, we can validate our hypothesis that VLMs perform well at VCU problems and LLMs help VCI problems better. We observe that, in VCU problems, the VLMs perform significantly better than LLM reasoning based on the caption on both datasets, with an average accuracy of 63.6\% vs. 56.0\%. While on VCI problems, LLM+caption performs better on average at 53.6\% vs. 50.5\%. We could also observe that BLIP2 has a significant performance gap between the two kinds of problems while LLM performs similarly. 

\vspace{1mm}
\noindent\textbf{How do visual factors and LLM clue reasoning help visual commonsense reasoning?} 
We validate the effectiveness of visual factors reasoning and LLM clue reasoning on both BLIP2-Pretrain and LLM-based decision paradigms. Here, we describe how we adapt the clue generation method (as in Eq.~\ref{eq:7}) for BLIP2-Pretrain decision paradigm: we first prompt the LLM to generate the required visual factors $f_j$, then generate visual clues $o_{ij}$ of these factors that can support each choice $i$. When applying the clues to BLIP2-Pretrain, we take the average of the image-text alignment scores within the same choice as the image-text alignment score for the choice $i$:
\begin{equation}
    S_i = \frac{1}{n}\sum_j (ITM(I, o_{ij}) + ITC(I, o_{ij}))
    \label{eq:10}
\end{equation}
where $n$ is the number of required visual factors determined by LLM. 
The choice with the highest score will be selected. 

From Table~\ref{tab:ablations}, we can first find that visual factors and visual clues are less helpful in \textbf{VCU} problems. On VCU problems, besides directly taking the concept being asked by the original question as the visual factor. The model will also consider low-level visual features as visual factors for the question. For example, for the question \texttt{What is the event in the image}, and the choice \texttt{dinner}, the visual factor could be \texttt{objects in the image}, and the reasoned visual clues could be \texttt{plates with food on the table}.

On BLIP2-Pretrain, using clues for image-text alignment is not better than using directly transferred declarative sentences. This validates that BLIP2 can already align visual features with different concepts well. 
However, introducing visual factors and observations as extra context improves performance on LLM reasoning, especially when the LLM is not confident about its initial judgment with only caption as context. In this case, the performance of LLM reasoning (`Cap + Clue' in Table~\ref{tab:ablations}) is comparable with pre-trained BLIP2. 

For \textbf{VCI} problems, visual factors and visual clue generations help both reasoning paradigms. 
First, the improvement in the BLIP2-Pretrain paradigm validates that \textbf{(1)} pre-trained BLIP2 cannot well-align statements that go beyond literal visual content, requiring commonsense inference; \textbf{(2)} LLM can reason about the visual factors that may contribute to supporting candidate commonsense inferences, and guide the VLM to focus on relevant factors accordingly.
Second, the improvement in the LLM reasoning paradigm shows that LLM clues successfully provide subtle details of the scene that are crucial for solving the problem.
Third, visual clues reasoned by LLM are better than VQA as the visual information provider. There are mainly two reasons. First, the pre-trained VLM sometimes could not understand or correctly answer the question due to the lack of language alignment. Second, the VQA model lacks the main question as the context and may not get the intention of the visual factor. Therefore, it may produce irrelevant answers. We provide examples to further illustrate these in Section~\ref{sec:6.3}.


\vspace{1mm}
\noindent\textbf{How to determine the reasoning process based on confidence and problem category for better LLM and VLM collaboration?} When deciding the reasoning process, we need to consider both the performance and efficiency, evaluating by the number of LLM calls. 
From Table.~\ref{tab:ablations}, we can observe that when the LLM is confident about its initial reasoning, the performance is the best or almost the best on both VCU and VCI problems. Therefore, using LLM+caption is the best choice. 
When the LLM is not confident about its initial reasoning on VCI problems, LLM+Caption+LLM clue significantly outperforms other decision paradigms. On VCU problems, we can observe that the performance of BLIP2 is similar to LLM+Caption+LLM clue. However, the LLM+Caption+LLM clue requires five LLM calls, which is three times more than using BLIP2. Therefore, using BLIP2-ITA is the best choice in this case.

\begin{figure*}[t]
\vspace{-4mm}
  \centering
  \includegraphics[width=0.82\textwidth]{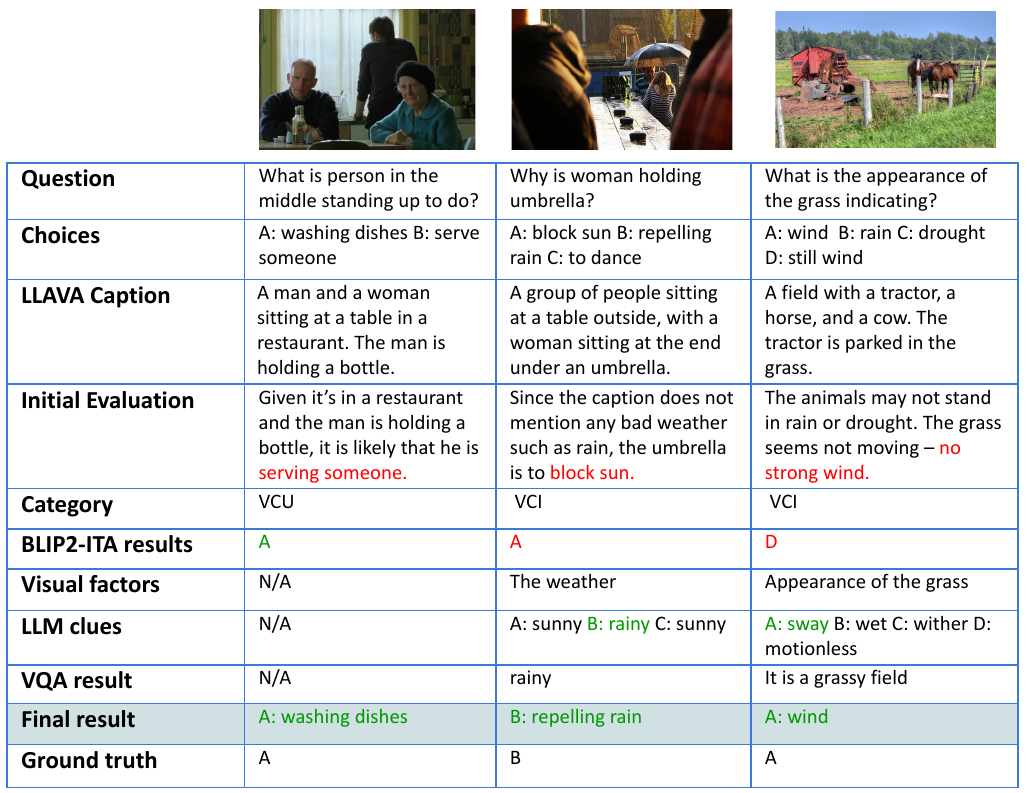}
  \vspace{-2mm}
  \caption{\textbf{Qualitative examples.} All the examples are in the case of initial reasonings are not confident. \textbf{Left:} An example in the \textbf{VCR} dataset, where the ITA corrects the initial reasoning. \textbf{Middle:} An example in the \textbf{A-OKVQA} dataset, where the LLM corrects the initial reasoning after giving the observation of the visual factor.  \textbf{Right:} An example in the \textbf{A-OKVQA} dataset, where the reasoned clue provides more useful information than VQA.}
  \label{fig:4}
  \vspace{-4mm}
\end{figure*}

\vspace{-2mm}
\subsection{Main Results}

\paragraph{VCR} The results on VCR dataset are in Table~\ref{tab:main vcr}. Our method achieves the best result compared with other training-free methods. Specifically, our method outperforms IdealGPT~\citep{you2023idealgpt} since it is able to leverage the visual understanding abilities of VLMs more effectively by considering the types and definitions of problems. However, we notice that there is still a significant gap between ICL methods and methods with supervised training. This could be due to the loss of information in approximating the naming and labeling of the persons mentioned in Section~\ref{sec:5.1}. 

\paragraph{A-OKVQA} On A-OKVQA dataset, on both GPT models, our method can improve on chain-of-thought baseline by a significant margin. Compared with concurrent method AssistGPT~\citep{gao2023assistgpt}, which utilizes GPT4 to call more visual tools such as object detection~\citep{liu2023grounding}, text detection, and region grounding~\citep{wang2022ofa}, our method with only BLIP2 and LLAVA can achieve better results. Meanwhile, we can observe that our method ViCor, without any training on the dataset, can achieve results close to the best supervised methods. This shows that our analysis and modeling for visual commonsense reasoning makes our framework tackle the VCR problems more efficiently. 

 \vspace{-1.5mm}
\subsection{Qualitative Examples} \label{sec:6.3}

In Fig.~\ref{fig:4}, we demonstrate several qualitative examples. The left example shows a case where the problem is classified as VCU, and the BLIP2-Pretrain selects the correct answer. 
The middle example presents a case where the initial evaluation is incorrect, and both the VQA and clue reasoning methods give the correct observation for the visual factor `weather', based on which the LLM selects the correct answer. The BLIP2-Pretrain here selects `block sun' due to the lighting condition of the image.
The example on the right demonstrates a case when the LLM reasoned answer is better than the answer generated by the VQA model. Here, the VQA does not understand the intention of the visual factor without the context of the main question. The LLM reasoned answer, however, can provide the most relevant information to the question and help the final reasoning. The BLIP2-Pretrain fails here due to the textual similarity between `wind' and `still wind'. 

\section{Conclusion}

In this work, we study the collaboration of pre-trained vision-language models and large-language models on a complex problem -- visual commonsense reasoning (VCR). 
We analyze and validate the distinct advantages of LLMs and VLMs by testing them on two different types of VCR problems. 
Based on this, we propose the ViCor framework that efficiently uses the visual understanding capabilities of VLMs and commonsense reasoning capabilities of LLMs to overcome the challenges in VCR. 
The experiment results validate the effectiveness of our framework. We believe our study can provide insights into the roles and the collaboration of LLMs and VLMs in vision and language problems. 

\section{Limitation and Potential Risk}

In our framework, we use text as the communication medium between LLMs and VLMs. The loss of visual details caused by captions may be hindering certain scenarios, and thus, our method lags behind supervised best-performing methods. Future work could explore incorporating our designs into an end-to-end fine-tuning approach. Large language models (LLMs) are a core component of our framework. Therefore, our method may inherit the potential risks from LLMs, such as hallucination and potentially offensive language. 

\bibliography{custom}

\clearpage

\appendix

\section{{Additional Results}} \label{appendix:a}

\subsection{ Results on More LLM Decoding Configurations}

\begin{table*}[t]
\centering
\setlength{\abovecaptionskip}{8pt}
\setlength{\belowcaptionskip}{8pt}
\caption{ Ablations on VCR with more decoding configurations.
}
\resizebox{0.8\textwidth}{!}{
\begin{tabular}{cccccc}
\toprule
\multirow{2}{*}{\textbf{Decision Model}}& \multirow{2}{*}{\textbf{Decoding Config}}  & \multicolumn{2}{c}{\textbf{VCU}} & \multicolumn{2}{c}{\textbf{VCI}}\\
    &   & Conf & !Conf & Conf & !Conf \\
  \midrule
 \multirow{4}{*}{LLM + Caption + LLM Clue} & Orig & 72.9 & 58.1 & 57.1 & 52.9  \\
& Temp 0.1 & 72.4 & 58.8 & 57.1 & 53.9 \\
& Temp 0.2 &  72.4 & 58.5  & 61.2 & 52.2 \\
& ICL examples &  74.7 & 56.7  & 59.2 & 54.2 \\
  \midrule
\multicolumn{2}{c}{Num. of Examples} & 170 & 1779 & 49 & 1002\\
\bottomrule
\end{tabular}
}
\label{tab:full vcr ablations}
\end{table*}
 To validate the robustness of our method, we ran the experiments on the VCR dataset with more decoding configurations using LLM + Caption + LLM Clue decision branch. Specifically, we ran on two more LLM decoding temperatures 0.1 and 0.2, and used different in-context examples for the prompt in Fig.3 (right) to guide the LLM to think about observations for visual factors based on candidate choices. From the results in Table.~\ref{tab:full vcr ablations}, we can observe that different decoding configurations influence the results by a small margin and do not affect the main conclusions.

\subsection{ Results on OKVQA dataset} 
 We adapt our method and baselines to OKVQA~\cite{okvqa} dataset. We use the full validation set, which contains 5046 examples. The results are in Table.~\ref{tab:okvqa}. We use GPT-3.5-Turbo for LLM modules. Since OKVQA is an open-ended dataset, we use the Caption+VQA clue version of our method in Figure 1 to tackle unconfident VCI problems. As shown above, our framework can still leverage the advantage of both VLMs and LLMs to achieve better results thanks to the better collaboration between them, e.g., VLMs utilization based on problem classification and active visual information acquisition. 
\begin{table*}[t]
\centering
\setlength{\abovecaptionskip}{8pt}
\setlength{\belowcaptionskip}{8pt}
\caption{ The result of ViCor on OKVQA dataset.
}
\resizebox{0.35\textwidth}{!}{
\begin{tabular}{cc}
\toprule
Method  & Accuracy \\
  \midrule
  LLM+Caption & 34.6  \\
BLIP2-T5XL &36.2  \\
 ViCor (ours) &  \textbf{38.7}  \\
\bottomrule
\end{tabular}
}
\label{tab:okvqa}
\end{table*}

\end{document}